\RequirePackage{fix-cm}
\documentclass[twocolumn]{svjour3}          
\smartqed  
\usepackage{graphicx}
\usepackage{hyperref}
\newcommand{\argmax}{\mathop{\mathrm{arg\,max}}}

\begin{document}

\title{An Integration of Bottom-up and Top-Down Salient Cues on RGB-D Data:
\thanks{This paper is based on the results obtained from a project commissioned by the New Energy and Industrial Technology Development Organization (NEDO), Japan.}
}
\subtitle{Saliency from Objectness vs. Non-Objectness}


\author{Nevrez Imamoglu \and Wataru Shimoda \and Chi Zhang \and Yuming Fang \and Asako Kanezaki \and Keiji Yanai \and Yoshifumi Nishida }


\institute{N. Imamoglu, A. Kanezaki, and Y. Nishida\at
              National Institute of advanced Industrial Science and Technology, Tokyo, Japan \\
              \email{nevrez.imamoglu@aist.go.jp}           
           \and
           C. Zhang and Y. Fang \at
               Jiangxi University of Finance and Economics, Nanchang, China
		   \and
           W. Shimoda and K. Yanai \at
               The University of Electro-Communications, Tokyo, Japan
            \and
}

\date{Preprint. This work includes the accepted version content of the paper published in \\Signal Image and Video Processing (SIViP), Springer,  Vol. 12, Issue 2, pp 307-314, Feb 2018.\\
DOI   \url{https://doi.org/10.1007/s11760-017-1159-7} }

\maketitle

\begin{abstract}
Bottom-up and top-down visual cues are two types of information that helps the visual saliency models. These salient cues can be from spatial distributions of the features (space-based saliency) or contextual / task-dependent features (object based saliency). Saliency models generally incorporate salient cues either in bottom-up or top-down norm separately. In this work, we combine bottom-up and top-down cues from both space and object based salient features on RGB-D data. In addition, we also investigated the ability of various pre-trained convolutional neural networks for extracting top-down saliency on color images based on the object dependent feature activation. We demonstrate that combining salient features from color and dept through bottom-up and top-down methods gives significant improvement on the salient object detection with space based and object based salient cues. RGB-D saliency integration framework yields promising results compared with the several state-of-the-art-models.
\keywords{Salient object detection \and Multi-model saliency \and  Saliency from objectness}
\end{abstract}

\section{Introduction}
\label{sec:intro}
Visual attention is an important mechanism of the human visual system that assists visual tasks by leading our attention or finding relevant features from significant visual cues \cite{1,2,3,4}. Perceptual information can be classified as bottom-up (unsupervised) and top-down (supervised or prior knowledge) visual cues. These salient cues can be from spatial distributions of the features (space-based saliency) or contextual / task dependent features (object based saliency) \cite{1,2,3,4}. 

Many researches have been done on computing saliency maps for image or video analysis \cite{5,6,7,8,9,10}. Saliency detection models in the literature generally demonstrate computational approaches either in bottom-up or top-down separately without integration of spatial and object based saliency information from image and 3D. Therefore, in this work, we introduce a multi-modal salient object detection framework that combines bottom-up and top-down information from both space and object based salient features on RGB-D data (Fig.1). 

\begin{figure*}
\begin{center}
  \includegraphics[width=0.95\textwidth]{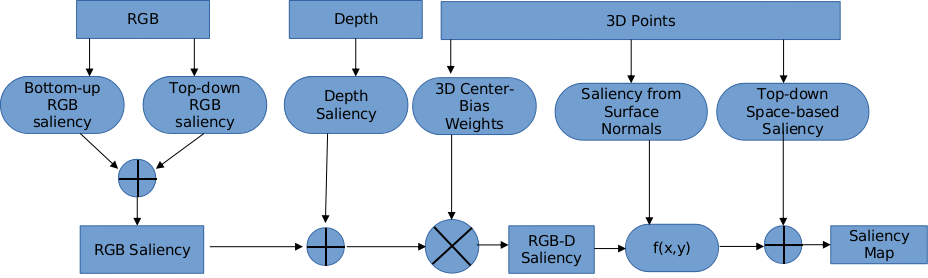}
\end{center}
\caption{Flowchart of the proposed saliency computation for integrating various perceptual salient cues}
\label{fig:1}       
\end{figure*}

In addition, regarding top-down saliency computation on color-images, we investigate salient object detection capability of various Convolutional Neural Networks (CNNs) trained for object classification or semantic object segmentation on large-scale data. Unlike the state-of-the-art deep-learning approaches to achieve salient object detection, we simply take advantage of the pre-trained CNN without the need of additional supervision to regress CNN features to ground truth saliency maps. The assumption is that prior-knowledge of CNNs on known objects can help us to detect salient objects, which are not included as trained object classes of the networks. We demonstrate that this can be done by object dependent features from both objectness and non-objectness scores of the CNNs. 

Finally, we examined integration of salient cues from depth and 3D point cloud information based on the spatial distribution of the observed scene in space. First, we use patch similarity based saliency computation on depth image for combining salient cues from color and depth images. Then, we apply weighting operation on color-depth salient cues based on two information: i) the distributions of normal vectors, and ii) center-bias weighting from joint 3D distribution of the observed scene as an improvement to the separate calculation on 2D image and depth demonstrated in \cite{21}. Finally, as top-down space-based saliency \cite{23}, spatial working memory of the robot is also included to support the attention cues based on the changes detected in the environment. 

In summary, combining color and depth information through bottom-up and top-down processes yielded significant improvement on salient object detection task from both space based and object based approaches. Our experimental evaluations on two different data-set from \cite{23} and  \cite{22} demonstrate promising results compared with the several state-of-the-art-models.

\section{Related Works}
\label{sec:relatedWorks}

Inspired by the studies on attention mechanism such as \cite{3}, first bottom-up (unsupervised) computational model of saliency was developed by Itti et. al \cite{6,4}, which was computing center-surround differences in multi-scale pyramid structure of images processed with various filters. Many researches followed the path of \cite{3} for computing saliency maps with bottom-up and top-down approaches on color and depth images \cite{4,5,6,7,8,9,10,21,23,22,24,35,36,37,38,39,40,11,12,13,14}. Among these models, bottom-up approaches rely on finding center-surround differences or contrast based on pixel or patch similarities such as WT \cite{24}, CA \cite{35}, SF \cite{37}, PCA \cite{38}, Y2D \cite{39,21}. Some of the works applied prior knowledge to obtain improved saliency results such as center-bias weighting Y3D \cite{39,21,23} or foreground/background location prior on super-pixels such as MR \cite{36}, RBD \cite{40}, RGBD \cite{22}.

Top-down saliency on color images have been explored in many works by taking advantage of bottom-up low-level features and/or high-level top-down features as training data. Supervised approaches on training data are applied to regress these extracted features to the ground truth salient object regions \cite{7,8,9,10}. After the recent developments and success of deep networks on image classification or segmentation tasks \cite{15,16,17,18}, similar methods were also implemented on features of CNNs trained for object detection or classification tasks as in \cite{11,12,13,14}. For example, MDF \cite{14} extracts multi-scale features from a deep CNN trained for large-scale object classification, then these features are used in an end-to-end (pixel-level) training with ground-truth saliency maps to yield saliency predictions for each pixel. However, these processes on trained CNNs require fine-tuning or another training to regress CNN features to infer saliency maps \cite{11,12,13,14}. 

Object or class dependent top-down saliency maps from color images can be obtained from objectness by using backward propagation on weakly supervised CNN models \cite{19,20}, which are generally trained for recognizing objects on the scene. In addition to the weakly supervised approaches, fully supervised models \cite{17,18} (i.e. fully convolutional neural networks) for semantic image segmentation result in several score maps to show the likelihood of representing each object class in different score or activation maps. However, all these models \cite{17,18,19,20} give likelihood maps or score maps for each class separately as CNN outputs. Therefore, in this work, we extend these works simply by combining these objectness based maps to one saliency map to investigate how these simple pooling handles general salient object detection tasks. In addition to objectness based saliency, we also show that score values for the background scene can aid salient object detection by simply using the negative likelihood of the background data as non-objectness map of CNNs. Our experiments demonstrate that these feed-forward computations from fully supervised models \cite{17,18} or backward process (gradients from back-propagation for CNN layers) from weakly supervised models \cite{19,20} can be representative of salient features. The intuition is that knowledge on the learned objects or learned features from CNNs can help to detect salient objects in the scene despite being unknown by the CNN as an object class. In contrast to the works in \cite{11,12,13,14}, the advantage of proposed simple usage of likelihood maps to obtain saliency is not having any additional training to regress CNN features to ground truth saliency maps.

Saliency features from depth \cite{21,22,39} or 3D \cite{21,23} information can improve the detection of salient cues on the perceived scene in addition to the color saliency. Saliency of depth images can be obtained similar to the color image saliency computation \cite{22,39}. For example, center bias-weighting can be applied to improve RGB-D saliency computation by applying center-bias on 2D image and depth \cite{21} separately as in \cite{39}. However, this can also be done in 3D jointly so we implemented center-bias weighting from joint 3D distribution of the observed scene to enhance saliency map computation. If there is a prior information on the spatial distribution of the scene as mean of the memory of the observer, detected changes in the environment can also affect the attention cues such as new objects in the scene, positional changes of the objects, and etc. \cite{29,30,31}. However, due to difficulty of obtaining data that enable prior information on the scene, it is not very common to take advantage of top-down space-based saliency based on finding the spatial changes in the environment as in \cite{23}. It is possible if only the observer has the memory of the observed scene so salient cues can be obtained through visual working memory by finding the changes in the spatial arrangements. If there is an object as a change in the environment, space-based saliency will be enough to detect salient object. However, if there are multiple new objects, new spatial arrangement of the environment, sensory error during the computation of changes, or etc., other salient cues will also be crucial to obtain reliable salient object detection results.

\section{Multi-model salient object detection}
\label{sec:proposed_framework}

Multi-modal saliency integration (Fig.\ref{fig:1}) consists of fusion of attention cues in RGB, Depth, and 3D data. Proposed saliency integration can be given as:

\begin{equation}
\label{eq:Eq1}
    \mathbf{S} = \mathrm{F}\left( \mathbf{S}_{RGBD}  +  \mathbf{S}_{SbS} \right)   
\end{equation}

\begin{equation}
\label{eq:Eq2}
   \mathbf{S}_{RGBD} = \mathrm{F}\left(\left(  \alpha\times\mathbf{S}_{RGB} +  
                                        \left(1-\alpha\right)\times\mathbf{S}_{D}\right)
                        \times\mathbf{W}_{cb}\right)^{\mathbf{S}_{N}}
\end{equation}

\begin{equation}
\label{eq:Eq3}
    \mathbf{S}_{RGB} =   \alpha\times\mathbf{S}_{RGB}^{TD} + \left(1-\alpha\right)\times\mathbf{S}_{RGB}^{BU}
\end{equation}

$\mathbf{S}$ is the proposed saliency map, $\mathbf{S}_{RGBD}$ is the color saliency ($\mathbf{S}_{RGB}$) and bottom-up depth saliency ($\mathbf{S}_{D}$) integration obtained as in Eq.\ref{eq:Eq2}, $\mathbf{w}_{cb}$ is the 3D center-bias weighting, $\mathbf{S}_{N}$ is salient cues obtained from the statistical distribution of the surface normal calculated for each image pixel, $\mathbf{S}_{SbS}$ is the top-down space-based saliency that requires visual memory of the environment, $\mathrm{F}\left( . \right)$ is the normalization function to scale the saliency values within {0-1} range, $\alpha= 0.7$ is empirically selected for weighted averaging process to give top-down salient cues more impact on the integration through averaging with a bottom-up saliency model. Saliency of color images ($\mathbf{S}_{RGB}$ in Eq.\ref{eq:Eq2} and Eq.\ref{eq:Eq3}) is obtained by two types of information; i) top-down RGB saliency ($\mathbf{S}_{RGB}^{TD}$), and ii) bottom-up RGB saliency ($\mathbf{S}_{RGB}^{BU}$).

\subsection{Saliency from color images}

Proposed computation of top-down color saliency maps and explanation of the bottom-up saliency model used in this paper (see Eq.\ref{eq:Eq3}) will be explained in the following sections.

\subsubsection{Top-down image saliency from objectness and non-objectness in CNN models}
\label{sec:sal_rgb_td}

Weakly or fully supervised convolutional neural networks (CNNs) can handle the process of salient object detection on a given scene even though given network may not be trained to classify the objects in the scene. However, feature representations of learned objects from the training data-set can help to represent similar objects on saliency detection task whether the CNN model knows the object/s of the scene or not. Then, we investigated efficiency of weakly and fully supervised models by introducing simple top-down saliency map computations using objectness or non-objectness values from pre-trained networks in the literature.

\paragraph{Saliency computation with fully supervised CNNs through objectness and non-objectness: }
\label{sec:sal_rgb_fully_objectness_non-objectness}

Two fully supervised CNNs (DeconvNet \cite{17} and SegNet \cite{18}) are tried to obtain top-down color image saliency, in which both CNNs are trained with PASCAL VOC data-set \cite{25} for semantic segmentation of 20 objects. Both models give object likelihood values at pixel level for each of 20 classes (i.e. objectness maps) and 1 likelihood map for the background class (i.e. non-objectness). In this part, we will demonstrate that top-down saliency from color images can be generated by both of these objectness or non-objectness likelihood maps. By using the objectness map from the DeconvNet \cite{17} and SegNet \cite{18}, salient features can be obtained as below: 

\begin{equation}
\label{eq:Eq4}
	\mathbf{S}_{RGB}^{TD} = \mathrm{F}\left( \frac{1}{C} \times 
	                                                 \sum_{c=1}^{C} \mathbf{O}_{c}\right)^{\mathbf{\lambda}}
\end{equation}

\begin{equation}
\label{eq:Eq5}
	\mathbf{\lambda} = \mathrm{F}\left( 
	                   \argmax_{c} \left( \left( \frac{\mathbf{O}_{c}-\mathbf{\mu}_{c}}
	                   {\mathbf{\sigma}_{c}} \right)^{2}\right)\right)^{\frac{1}{2}}
\end{equation}

In Eq.\ref{eq:Eq4}, \textit{c} is the object or class index, \textit{C} is the number of objects/classes that can be recognized by the CNN model, $\mathbf{O}_{c}$ is the score map for each class \textit{c}, $\mathbf{\lambda}$ (given in Eq.\ref{eq:Eq5}) is the parameter for amplifying by taking the maximum of the normalized objectness value (see Eq.\ref{eq:Eq5}) over all classes, $\mathbf{\mu}_{c}$ is the mean objectness of the class c, $\mathbf{\sigma}_{c}$ is the standard deviation of $\mathbf{O}_{c}$ for each object score maps. 

In addition to the saliency through objectness, we also create saliency by using the negative non-objectness likelihood map in a very simple way. Again, we use both DeconvNet \cite{17} and SegNet \cite{18} to provide non-objectness likelihood map. Top-down saliency from non-objectness also gives quite good results even though it is not the best (see experimental results section). So, non-objectness based top-down salient features can be given as below,where $\mathbf{NO}_{c}$ is the non-objectness likelihood map scaled to {0-1} range to obtain, $\mathbf{S}_{RGB}^{TD}$, top-down object saliency through non-objectness.

\begin{equation}
\label{eq:Eq6}
	\mathbf{v}_{i}^{c} = \mathrm{F_{scale}}\left( \mathbf{-NO}_{c}\right)
\end{equation}

\paragraph{Saliency computation with weakly CNNs through objectness: }
\label{sec:sal_rgb_weakly_objectness}

One of the main problems of fully supervised CNN models as stated in \cite{20} is to require a large set of training data since ground truth labels should be assigned for each pixel of each image in the training data-set. Therefore, saliency computations using weakly supervised models are demanded, and have big advantage on creating large training data. Unlike DeconvNet or SegNet with only 20 classes of object recognition capability, it is easier to create large training data for weakly supervised models with many object category such as ImageNet with 1000 object class. 

Saliency through back-propagation process in \cite{19,20} demonstrated that regions which have high score derivatives respond to object location. In other words, these regions are related to object based attention cues. Therefore, we use VGG16 CNN model trained with ImageNet by Simonyan et al. \cite{15,19}. Simonyan et al. \cite{19} regarded the derivatives of the class score with respect to the input image as class saliency maps. However, we use the derivatives of relatively upper intermediate layers which are expected to retain more high-level semantic information by extending our previous work from Shimoda et al. \cite{20}. The difference is to obtain saliency for all objects in one saliency map rather than computing class level saliency map for each object class category. Finally, we average them to obtain one saliency map. The procedure can be given as follows \cite{20,19}: 

\begin{equation}
\label{eq:Eq7}
	\mathbf{v}_{i}^{c} =\frac{\partial\mathbf{S}_{c}}{\partial\mathbf{L}_{i}}
\end{equation}

\begin{equation}
\label{eq:Eq8}
	\mathit{m}_{i,x,y}^{c} = \mathsf{max}  \left(   \mathbf{w}_{i,\mathit{h}_{i}\left( x,y,k\right)}^{c} \right)
\end{equation}
 
\begin{equation}
\label{eq:Eq9}
	\mathit{g}_{x,y}^{c} = \frac{1}{L}\sum_{i=1}^{L} \tanh (a\cdot\mathit{m}_{i,x,y}^{c})
\end{equation}

\begin{equation}
\label{eq:Eq10}
	\mathbf{S}_{TD}^{RGB} =\frac{1}{K}\sum_{c=1}^{K} \mathit{g}_{x,y}^{c}
\end{equation}

The class score derivative $\mathbf{v}_{i}^{c}$ (Eq.\ref{eq:Eq7}) of a feature map of the $\mathit{i}^{th}$ layer (\textit{i}=\{3,4,5\}) is the derivative of class score with respect to the layer. $\mathbf{v}_{i}^{c}$ can be computed by guided back-propagation (GBP) \cite{26,20} instead of back propagation (BP). In the GBP, only positive loss values are propagated back to the previous layers through ReLUs. GBP can emphasize salient cues on the objects such as edges or boundaries of the objects along with the textures \cite{20}. After obtaining $\mathbf{v}_{i}^{c}$, we up-sample it to $\mathbf{w}_{i}^{c}$ with bi-linear interpolation so that the size of a 2-D map of  becomes the same as an input image as in Eq.\ref{eq:Eq8}, where $\mathit{h}_{i}\left( x,y,k\right)$ is the index of the element of $\mathbf{w}_{i}^{c}$, $\mathit{k}$ represents kernel.  $\mathit{m}_{i,x,y}^{c}$ is aggregated for each target layer to obtain saliency feature maps $\mathit{g}_{x,y}^{c}$ as in  Eq.\ref{eq:Eq9}. In  Eq.\ref{eq:Eq9},  \textit{a} is a scaling factor defined as 3 as in \cite{20}, \textit{L} is the number of layer to aggregate. Then, in contrast to distinct class saliency maps \cite{20}, here, we introduce to combine all class dependent salient cues to obtain an objectness based final saliency map. For this reason, top \textit{k} = 3 class of recognition result as score maps are combined to obtain over all top-down saliency (Eq.\ref{eq:Eq10}).

\subsubsection{Bootom-up image saliency from low-level features}
\label{sec:sal_rgb_bu}

Saliency from low-level features of color images, we use our work in \cite{24} since this part is not the main contribution of this study. However, we demonstrate that including bottom-up salient cues can improve the saliency obtained from color images while combining with the top-down salient cues. This model uses Wavelet Transform (WT) \cite{24} to obtain local and global salient features. WT model \cite{24} only relies on bottom-up low-level features representing attention areas such as edges and color contrasts without any prior information or knowledge. But, any bottom-up model would be replaced in this framework to investigate which bottom-up and top-down models are more complementary to each other. 

\subsection{Salient cues from depth image and 3D data for RGB-D saliency}
\label{sec: sal_cues_depth_3d_bu}

\paragraph{Depth Saliency: }
\label{sec:sal_depth_bu}
Depth saliency, $\mathbf{S}_{D}$, is computed with same approach as described by Yuming et. al.\cite{21}. This approach takes advantage of comparing patch features with the surrounding patches by distance weighting, where model utilizes DCT based saliency computation. Similar to the bottom-up color saliency calculation of the same work.

Other than the color and depth saliency, we can take advantage of 3D spatial distribution of the scene for improving the saliency computation with additional information such as 3D center-bias weighting and salient cues from surface normal.

\paragraph{Center Bias Weighting: }
\label{sec:sal_3D_cbw}

Yuming et. al. \cite{21} showed that integration of center bias theorem to scene content can improve the performance of saliency maps. They used two types of center bias: i) on image space that applies center bias on pixel indexes regarding X and Y of the scene, and ii) on the range of depth values. However, if 3D data are available for each pixel, center bias weights can be obtained in 3D space jointly in a more uniform space rather than having center bias weighting on image and depth separately. Therefore, in this work, we utilized adaptive 3D center bias weighting depending on the visible depth information. The center bias weighting can be calculated as below: 

\begin{equation}
\label{eq:Eq13}
	\mathbf{W}_{CB} = e^{ \left(c_{h}\times \frac{\mathbf{h}}{2\sigma^2}  - c_{v}\times \frac{\mathbf{v}}{2\sigma^2} - \frac{\left( \mathbf{d} -\mathsf{min}(\mathbf{d}|\mathbf{d}>0) \right)}{2\sigma^2}                                                    \right)	}
\end{equation}

$\mathbf{W}_{CB}$ (Eq.\ref{eq:Eq13}) is the center-bias weight matrix. In Eq.\ref{eq:Eq13}, \textbf{h}, \textbf{v}, and \textbf{d} are 3D data matrix, \textbf{d} is depth, \textbf{h} is height or vertical axis, and \textbf{v} is for the width or horizontal axis. All zero values of \textbf{d} are assigned to minimum non-zero value in \textbf{d}. $\sigma$ is the deviation around the focus area regarding the center-bias, which is defined as $\sigma=\eta \times \mathsf{max}(\mathbf{d})$ , and $\eta$ is empirically assigned to 0.25.  $c_{v}$ and $c_{h}$ are the scaling constant on horizontal and vertical points to give more priority to depth on 3D center-bias weighting, and they are also used during normal calculations for smoothing the the data to avoid non-number values or noise since Kinect sensor can not measure close distance and it does not give accurate values at more than 5 meters depth measurement.

\paragraph{Weighting from the distribution of Surface Normal Vectors:}
\label{sec:sal_depth_surfacenormals}
After we obtain center-bias weighting to improve attention cues, we also find salient cues for 3D points by checking the distribution of the surface normal of each point. First, we calculate normal of each 3D point by creating a surface with the points within a defined radius \cite{27}. Then, inspired by \cite{28}, for each point represented with the surface normal, we calculate the Mahalanobis distance (Eq.\ref{eq:Eq15}) to the cluster of all 3D points (i.e. distribution of the normal vector of all 3D points).

\begin{equation}
\label{eq:Eq15}
	\mathit{S}_{N}=  \mathrm{F_{scale}}\left( \mathbf{n} \times \mathbf{\Sigma}^{-1} \times \mathbf{n}^\mathtt{T}\right)
\end{equation}

\textbf{n} is the 3D normal vectors of each point corresponding to each pixel at color image, $\Sigma$ is co-variance matrix of \textbf{n}. To remove the noise on the $\mathbf{S}_{N}$, we apply 2D median filter with \{5,5\} window after reshaping the $\mathbf{S}_{N}$ to 2D saliency map. Also, we do enhancement by increasing the saliency values of points around the peak salient cues with values higher than 0.8 [21, 24].

\begin{table*}[]
\centering\small\setlength\tabcolsep{.95em}
\caption{Evaluation of the objectness/non-objecteness color saliency maps using Area Under Curve (AUC) metric}
\label{tab:1}       
\begin{tabular}{llllllll}
\hline\noalign{\smallskip}
DO & DNO & SO & SNO & GBP &  \textbf{DOC} &  \textbf{SNOC} & \textbf{GBPC}\\
\noalign{\smallskip}\hline\noalign{\smallskip}
0.8962 & 0.8826 & 0.7424 & 0.9044 & 0.9256 & 0.9281 & 0.9219 & 0.9368 \\
\noalign{\smallskip}\hline
\end{tabular}
\end{table*}

\subsection{Top-down space-based saliency from 3D points}

We apply our previous work \cite{23} for top-down space-based saliency model relying on detecting changes in the environment from 3D Kinect observations. Data-set to create space-based saliency is obtained from a mobile robot monitoring system, which is established with Pioneer P3-DX mobile robot with a Laser Range Finder (LRF) for localization and mapping and a Kinect sensor on a rotating platform for observing the scene \cite{23}. Local Kinect point cloud scene is projected on the 2D global map (i.e. obstacles and free regions in the room). Global map as the spatial memory of the robot is created by using SLAM \cite{32,33} with LRF sensor \cite{23}. Using the changes in the environment based on the projected data and depth information, space-based attention map is obtained to be the top-down salient cues from 3D points (see \cite{23} for details). The performance improvement through the top-down space-based saliency are given in experimental results. 


On the other hand, change based saliency is only reliable when we have prior information of the environment as stated before. For example, if a mobile observer (e.g. robot) enters a new environment with no prior knowledge or if there is a sensory error in localization, change detection based saliency will not be reliable and change may not be the only needed attention cues. Hence, we need other salient cues to analyze the scene. In sum, to define focus of attention in the scene, we need to take advantage of a multi-model saliency computation framework as proposed in this work. These bottom-up and top-down salient cues includes features obtained from color image, depth image, and 3D data. In the following section, we demonstrate that our multi-modal salient feature fusion can give very reliable results for salient object detection.

\section{Experimental Results}
\label{sec:results}

Proposed multi-modal saliency framework is tested in two different data-sets. First one is  RGB-D saliency data-set (RGB-D-ECCV2014) used for evaluation of salient object detection work in \cite{22}. RGB-D-ECCV2014 data-set, 1000 color images and depth data, is used to compare our proposed saliency framework with the several state of the art models. However, in this comparison, we exclude change detection related top-down space-based saliency since RGB-D-ECCV2014 data-set does not have any information for any reference environment memory to aid comparison of current spatial observation with past state or spatial placement of the scene. For the second data-set (ROBOT-TCVA2015), we used the data from \cite{23} with RGB color images and 3D Kinect data, which is recorded from a mobile robot while a person was doing various indoor activities in a room. This data also includes 2D global map of the environment with robot pose and Kinect pose in the room for each frame. So, this information can help to project local Kinect data on global 2D map to detect possible attention changes in the room.

\paragraph{Evaluation of top-down saliency on color images:}
\label{sec:eval_rgb}

We introduce top-down image saliency computation from the fully-supervised (DeconvNet \cite{17} and SegNet \cite{18}) or weakly-supervised (VGG-16 \cite{26,20}) pre-trained models (see Section 3.1). In this section, we compare proposed DO and DNO (objectness and non-objectness based saliency using DeconvNet \cite{17}), SO and SNO (objectness and non-objectness based saliency using SegNet [18]), and GBP (objectness based saliency using Guided Back-propagation \cite{26,20}). Regarding the non-objectness based saliency using GBP, we tried to train the weakly supervised CNN model by including class output for the background or non-objectness class as in fully supervised models; however, we could not get reliable results from these training attempts. Therefore, we did not include trials for saliency from non-objectness and GBP combinations in weakly supervised models. 

Comparison of AUC results based on the ECCV2014 RGBD data-set are given in Table.\ref{tab:1}. An interesting observation is seen on the SegNet \cite{18}, in which non-objectness based saliency (SNO) outperformed objectness based saliency (SO) on SegNet \cite{18} model. Moreover, SO results in considerably poor results as a top-down saliency approach through objectness. This observation on SO performance shows that SegNet \cite{18} still open to improvement for object representation and classification, which can be achieved by introducing multi-task learning as a future work. Among these five proposed top-down color image saliency trials, GBP, SNO and DO had the top three AUC values. So, we will use these top three performing model for further evaluation while we apply our multi-model saliency framework. Then, for comparison, we combined each of the top-down saliency approaches GBP, SNO, and DO with bottom-up model to obtain color image saliency map $\mathbf{S}_{RGB}$ (Eq.3). We refer these color saliency map variations using GBP, SNO, and DO as GBPC, SNOC, and DOC, respectively. 

\begin{table*}
\centering\small\setlength\tabcolsep{.99em}
\caption{Area Under Curve (AUC) based evaluation of the selected models in the literature and our proposed saliency computations (SNOP, DOP, GBPP) within our multi-modal framework}
\label{tab:2}       
\begin{tabular}{lllllll}
\hline\noalign{\smallskip}
SF & WT & CA & Y2D & RGBD & PCA & Y3D \\
\noalign{\smallskip}\hline\noalign{\smallskip}
0.7637 & 0.8453 & 0.8488 & 0.8859 & 0.9033 & 0.9089 & 0.9094 \\
\hline\noalign{\smallskip}
\hline\noalign{\smallskip}
RBD & MR & MDF & \textbf{SNOP} & \textbf{DOP} & \textbf{GBPP}  & \\
\noalign{\smallskip}\hline\noalign{\smallskip}
0.9170 & 0.9283 & 0.9328 & \textbf{0.9339} & \textbf{0.9398} & \textbf{0.9491} & \\
\noalign{\smallskip}\hline
\end{tabular}
\end{table*}

\begin{figure*}
\begin{center}
  \includegraphics[width=0.95\textwidth]{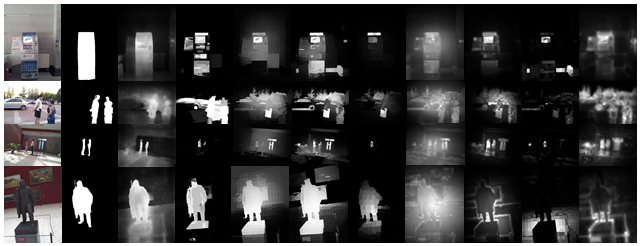}
\end{center}
\caption{ (a) sample color images with their (b) gorund truths, and saliency results of (c) our framework with GBPP, and other selected models (d) MDF \cite{14} (e) MR \cite{36} (f) RBD \cite{40} (g) RGBD \cite{22} (h) Y3D \cite{21} (i) PCA \cite{38} (j) SF \cite{37} (k) CA \cite{35}}
\label{fig:4}       
\end{figure*}

\paragraph{Comparison using RGB-D-ECCV2014 data-set:}
\label{sec:comp_eccvdata}

In this section, we will demonstrate the effectiveness of the proposed saliency model as an RGB-D saliency framework without the use of top-down space-based saliency (see Fig.\ref{fig:1}), when prior environment data (memory) is not available. We use Area Under Curve (AUC) measure obtained from the Receiver Operating Characteristic (ROC) curve \cite{34,24,21} as evaluation metric to compare the performances of the state-of-the-art models and proposed framework. We compared our results with various RGB and RGB-D based saliency models. The saliency works, WT \cite{24}, CA \cite{35}, MR \cite{36}, SF \cite{37}, PCA \cite{38}, Y2D \cite{39}, Y3D \cite{21}, RBD \cite{40}, RGBD \cite{22}, MDF \cite{14} are included in our comparisons.

On RGB-D-ECCV2014 data-set, it is clear that proposed RGB-D saliency performances improved while our GBPP, SNOP, and DOP (see Table.\ref{tab:2}) having AUC values 0.9491, 0.9339, and 0.9398, which are higher compared to their color only respective proposed variants (see Table.\ref{tab:1}). Among the state-of-the-art models, MDF \cite{14} has the best AUC performance, followed by MR \cite{36}. In summary, proposed models (SNOP, DOP, GBPP), outperformed the state-of the-art saliency models compared with. And in overall evaluation, our GBPP using weakly supervised CNN trained for 1000 object class has the best AUC performance on this data-set. In Fig.\ref{fig:4}, some color images and their corresponding saliency maps from the proposed GBPP and state-of-the-art models are given.

\paragraph{Comparison using ROBOT-TCVA2015 data-set:}
\label{sec:comp_robotdata}

We validated that proposed framework gives promising results on a RGB-D public data-set (RGB-D-ECCV2014 [22]) in previous experimental results. However, we were not able to use top-down space based saliency integration previously. Therefore, we will express the improvement of selective attention cues depending on the spatial changes in the environment.

We will use ROBOT-TCVA2015 data-set \cite{23} for this purpose, which consists of frames recorded in a room with a subject doing daily activities. The activities include tasks such as standing, sitting, walking, bending, using cycling machine, walking on treadmill, and lying-down, and etc. Since the environment is similar within an activity, we selected 10 frames randomly for representing all activities in these test samples. Then, we manually created ground truth binary images, where the subject is the focus of attention. Then, we tested proposed framework fully with the salient cues obtained from the changes in the environment. ROBOT-TCVA2015 includes Kinect data and global map with robot pose recorded for all frames. So, this allows us to create top-down space-based saliency by projecting the local Kinect data to global map to find changes and attention values to these changes \cite{23}.

\begin{figure*}[ht]
\begin{center}
  \includegraphics[width=0.925\textwidth]{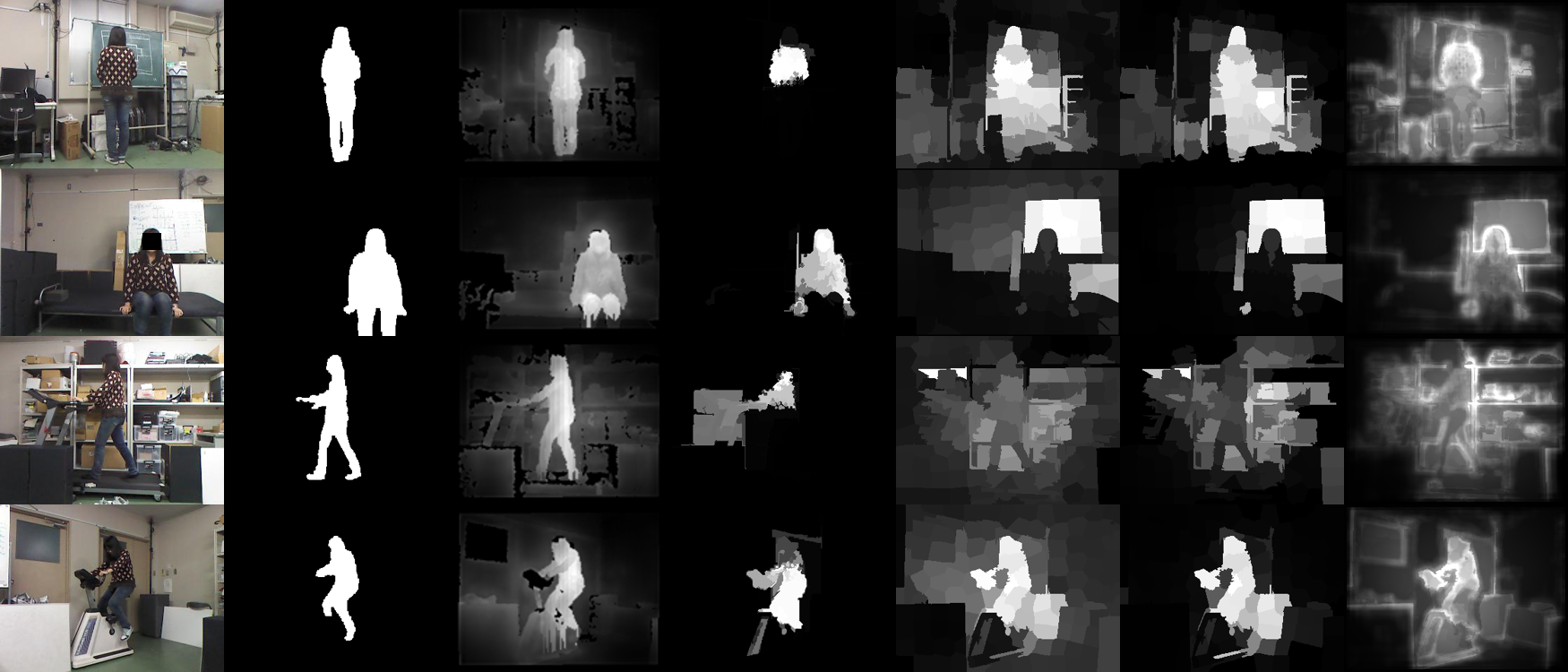}
\end{center}
\caption{Saliency results of (a) sample images with (b) ground truth using models: (c) our GBPP-SbS (d) MDF \cite{14} (e) MR \cite{36} (f) RBD \cite{40} (g) PCA \cite{38}}
\label{fig:6}       
\end{figure*}

\begin{table*}[ht]
\centering\small\setlength\tabcolsep{.95em}
\caption{Evaluation of the selected models and our GBPP-SbS using Area Under Curve (AUC) metric}
\label{tab:3}       
\begin{tabular}{llllll}
\hline\noalign{\smallskip}
MR & PCA & RBD & MDF & \textbf{GBPP} & \textbf{GBPP-SbS} \\
\noalign{\smallskip}\hline\noalign{\smallskip}
0.8060 & 0.8659 & 0.7518 & 0.8657 & \textbf{0.9468} & \textbf{0.9592} \\
\noalign{\smallskip}\hline
\end{tabular}
\end{table*}

Proposed framework (Fig.\ref{fig:1}) can be implemented fully by combining salient cues from color images (bottom-up and top-down saliency of color images), depth (bottom-up depth saliency), center-bias weighting, normal vector (saliency weighting from the distribution normal vectors), and spatial changes in the environment (top-down space based saliency) as resulting GBPP-SbS in this experiment. For comparison on ROBOT-TCVA2015 data, four best performing state-of-the-art models are selected from the previous analysis, which are MR \cite{36}, PCA \cite{38}, RBD \cite{40}, and MDF \cite{14} models. From our proposed variants, the best performing case, GBPP, is used to compare with other models and also to check the improvement when we combine top-down space-based saliency with GBPP which is labelled as GBPP-SbS. In Fig.\ref{fig:6}, some sample images, ground truth (GT) salient object (person in the data), and their corresponding saliency examples for some of the state-of-the-art models and our saliency framework (GBPP and GBPP-SbS) are given.

Our proposed  GBPP-SbS shows the best AUC performance (0.9592) for the ROBOT-TCVA2015 test data among the all compared models (see Table.\ref{tab:3}). In this test, AUC performances of the selected state-of-the-art models decrease drastically compared to the test results on RGB-D-ECCV2014 data-set in previous section. Perhaps, real-time data from an uncontrolled environment effected their accuracy on detecting salient cues due to noise and high illumination change conditions in ROBOT-TCVA2015 data. MR \cite{36}, PCA \cite{38}, RBD \cite{40}, and MDF \cite{14} have AUC performances as 0.8657, 0.8659, 0.7518, and 0.8060 respectively. On the other hand, top-down space-based saliency from detected changes improves the AUC performance of the GBPP from 0.9468 to 0.9592 for the proposed GBPP-SbS saliency maps.

\section{Conclusion}
\label{sec:conclusion}
Proposed work demonstrates a saliency framework that takes advantage of various attention cues from RGB-D data. The model demonstrated its reliability from two different data-sets compared to the state-of the-art models. However, even though saliency results on mobile robot data having promising performance, the current framework is not suitable for real-time computation. Therefore, we would like to extend and improve the model for real-time mobile robot surveillance. In summary, we applied proposed saliency integration framework step-by-step to obtain saliency on color images, then RGB-D, and finally RGB-D with top-down space based saliency. Evaluation from AUC metric shows importance of multi-model saliency from both spatial and object based salient cues. Especially, saliency analysis on CNNs from objectness and non-objectness shows interesting findings for these networks trained for object classification or segmentation. 

\begin{acknowledgements}
Dr. Nevrez Imamoglu thanks Dr. Boxin Shi at AIST (Tokyo, Japan) for discussions on 3D data processing.
\end{acknowledgements}



\end{document}